\documentclass[runningheads]{llncs}
\usepackage[T1]{fontenc}
\usepackage{url}
\usepackage[utf8]{inputenc}
\usepackage{float}

\usepackage{hyperref}
\usepackage{amssymb}
\usepackage{array}
\usepackage{tabularx}
\usepackage{pifont}
\usepackage{multirow}

\usepackage{pifont} 
\usepackage{geometry} 
\usepackage{lipsum}
\usepackage{colortbl}

\usepackage{graphicx}
\usepackage{orcidlink} 

\begin{document}
\title{Fruit Deformity Classification through Single-Input and Multi-Input Architectures based on CNN Models using Real and Synthetic Images}
\titlerunning{Fruit Deformity Classification}
%
\author{
  Tommy D. Beltran\inst{1}\orcidlink{0009-0001-0150-7474} \and
  Raul J. Villao\inst{1}\orcidlink{0009-0009-0610-412X} \and
  Luis E. Chuquimarca\inst{1,2}\orcidlink{0000-0003-3296-4309} \and
  Boris X. Vintimilla\inst{1}\orcidlink{0000-0001-8904-0209} \and
  Sergio A. Velastin\inst{3,4}\orcidlink{0000-0001-6775-7137}
}
\authorrunning{T. D. Beltran et al.}
%
\institute{
  \inst{1}ESPOL Polytechnic University, CIDIS, Guayaquil, Ecuador\\
  \inst{2}UPSE Santa Elena Peninsula State University, FACSISTEL, La Libertad, Ecuador\\
  \inst{3}Queen Mary University of London, School of EECS, London, UK\\
  \inst{4}University Carlos III of Madrid, Dept. of Computer Science, Madrid, Spain\\
  \email{\{tomdbelt, rjvillao, lchuquim, boris.vintimilla\}@espol.edu.ec}, \email{sergio.velastin@ieee.org}
}
%
\maketitle
\thispagestyle{plain}
\footnotetext{
Springer Copyright Notice \\
Copyright (c) 2024 Springer \\
This work is subject to copyright. All rights are reserved by the Publisher, whether the whole or part of the material is concerned, specifically the rights of translation, reprinting, reuse of illustrations, recitation, broadcasting, reproduction on microfilms or in any other physical way, and transmission or information storage and retrieval, electronic adaptation, computer software, or by similar or dissimilar methodology now known or hereafter developed. \\
Accepted to be published in: 2024 27th Iberoamerican Congress on Pattern Recognition (CIARP’24), Nov 26–29, 2024. This is an author-prepared version. The final publication is available on Springer (DOI: \href{https://doi.org/10.1007/978-3-031-76607-7_4}{$10.1007/978-3-031-76607-7_4$}).
}

\begin{abstract}
The present study focuses on detecting the degree of deformity in fruits such as apples, mangoes, and strawberries during the process of inspecting their external quality, employing Single-Input and Multi-Input architectures based on convolutional neural network (CNN) models using sets of real and synthetic images. The datasets are segmented using the Segment Anything Model (SAM), which provides the silhouette of the fruits. Regarding the single-input architecture, the evaluation of the CNN models is performed only with real images, but a methodology is proposed to improve these results using a pre-trained model with synthetic images. In the Multi-Input architecture, branches with RGB images and fruit silhouettes are implemented as inputs for evaluating CNN models such as VGG16, MobileNetV2, and CIDIS. However, the results revealed that the Multi-Input architecture with the MobileNetV2 model was the most effective in identifying deformities in the fruits, achieving accuracies of 90\%, 94\%, and 92\% for apples, mangoes, and strawberries, respectively. In conclusion, the Multi-Input architecture with the MobileNetV2 model is the most accurate for classifying levels of deformity in fruits.

\end{abstract}
\section{Introduction}
\label{sec:intro}

One of the most critical stages in the post-harvest process of fruits and vegetables is the selection and classification through quality inspection (parameters of ripeness, deformation, and defects)~\cite{olorunfemi2021post}. This process, typically meticulous and mechanized, is usually carried out manually by specialized personnel in the field. The significance of this stage lies in the commercial value attributed to the product once evaluated, as higher-quality products tend to command higher selling prices in the market. Additionally, both small-scale retailers and supermarkets, driven by their commercial needs, also consider various parameters when purchasing these foods~\cite{shewfelt2022challenges}. Moreover, consumers often take into account the visual appeal of the fruit when making their purchases.

The issue lies in the fact that the quality assessment of fruits and vegetables is carried out through manual processes, turning it into a slow and inconsistent procedure due to the subjectivity associated with the evaluation criteria of the individuals responsible for inspection. This human factor diminishes the efficiency of the evaluation process and adversely impacts the commercial value of the fruits, affecting both consumers in general and farmers/suppliers, who may experience decreased profits due to the quality of their products. Therefore, the purpose of this study is to propose an improvement in the fruit classification process by developing deep learning models that enable the automatic classification of one of the main quality assessment parameters: the level of fruit deformation. In this work, apples, mangoes, and strawberries are analyzed as the fruits under study.

This article is structured as follows. Section \ref{sec:literature review} provides a review of the state of the art regarding CNN models applied to the identification of fruit deformities. Section \ref{sec:methodology} describes the contributions and the proposed methodology for conducting the work. In Section \ref{sec:results}, the results of identifying levels of deformities in fruits using Single-Input and Multi-Input architectures with public CNN models are presented. Finally, the study's conclusions are provided in Section \ref{sec:conclusions}.

\section{Literature Review}
\label{sec:literature review}

This section describes the state of the art concerning the analysis of deformities in fruits, specifically apples, mangoes, and strawberries. It also reviews the techniques and models that have been employed in previous studies for deformity analysis in these fruits.

\subsection{Fruit Quality Inspection}

Quality inspection is an indispensable aspect of the food industry. Numerous food items and pre-processed products rely on fruits in their production, prompting both retailers and wholesalers to seek fruits of the highest quality and appearance. This approach ensures consumer satisfaction and may result in higher remuneration for the seller when marketing their products.

To assess fruit quality, two methodological approaches are employed: invasive and non-invasive methods~\cite{vetrekar2015non}. Invasive methods, also referred to as destructive, facilitate the analysis of attributes such as internal color, texture, and properties like sugar content, and acidity, among others, through fruit manipulation. On the other hand, non-invasive methods enable the evaluation of external features such as deformities (shape, size), ripeness level, and defect detection without compromising the fruit's integrity.

International markets demand that fruits undergo rigorous quality control \cite{chuquimarca2023banana}, leading to the establishment of international standards by organizations such as the Economic Co-operation and Development (OECD), United States Department of Agriculture (USDA), and the regulatory bodies of the People's Republic of China.

When it comes to deformities, it is crucial to assess both the shape and size of the fruit. While Chinese and American regulations address both characteristics, European standards focus solely on the fruit's shape. Incorporating the evaluation of both characteristics would ensure a more accurate prediction of deformities. However, considering size adds a level of complexity to data collection. Given the impracticability of establishing a controlled environment for image acquisition and the paucity of publicly available data concerning size evaluation, this project opted to focus solely on the parameter of shape rather than size. Consequently, only shape classes delimited by the standards set forth by the OECD for deformity assessment are taken into consideration. For this reason, the categories employed for each fruit were: Extra Class, First Class, Second Class, and Ungraded which will be the class where we will send all those images that are already outside of any limit allowed by the OECD.

\subsection{Fruit Deformities}

During the growth and development process of fruits, alterations in their shape and size can occur as a result of various factors, including genetic causes~\cite{behera2020identification}. A deformed fruit exhibits an atypical shape that does not align with the characteristic features of its species. In this study, we investigate the detection of deformities in fruits using CNNs. It is essential to consider the nature of these deformities to achieve precise classification and enable CNNs to learn to recognize fruits with deformities.

When dealing with deformities, typically two aspects are evaluated: shape and size. International organizations responsible for establishing quality parameters assess these aspects. Some examples include the OECD~\cite{lidror1990improving}, the USDA~\cite{huang2022us}, and governmental entities that establish their criteria, such as the Republic of China~\cite{zhou2015food}. Table \ref{tab:Parameters evaluated} summarizes the primary parameters evaluated in most fruits.

\begin{table}
\caption{Parameters evaluated for classifying deformities according to various international organizations.}
\label{tab:Parameters evaluated}
\centering
\begin{tabularx}{\textwidth}{|l|X|X|X|}
\hline
\textbf{Organizations} & \textbf{Categories} & \textbf{Shape} & \textbf{Size} \\ \hline
OECD & 
Extra-Class, Class 1, Class 2 &
\checkmark\ Typical Shape of fruit &
\ding{55} Does not assess \\ \hline

USDA & 
Extra-Fancy, Fancy, Utility &
\checkmark\ Typical Shape of fruit &
\checkmark\ Measurement of the fruit's major diameter. \\ \hline

Republic of China & 
Excellent, Class 1, Class 2, Substandard &
\checkmark\ Typical Shape of fruit &
\checkmark\ Measurement of the fruit's major diameter. Measurement of the fruit's circumference from the top view. \\ \hline
\end{tabularx}
\end{table}

In~\cite{wang2022grading} and~\cite{hu2021infield}, both the shape and diameter of the apple are considered for categorization. However, the limitation of size as an evaluation parameter is that it requires the measurement of fruit diameters. Therefore, it is necessary to capture images under controlled conditions, taking into account parameters such as the working distance between the camera and the target to facilitate diameter calculation.

According to the OECD, for a fruit to be classified as Extra Class, Class I, or Class II, it must meet minimum requirements for each parameter in quality inspection such as ripeness, defects, and shape. If a fruit fails to meet the criteria for classification within these classes, it is deemed of low quality and categorized as Ungraded. The following sections elaborate on the permissible levels of deformities in apples, mangoes, and strawberries within international standards for commercialization.

\subsubsection{Apple Deformities}

\begin{figure}
  \centering
   \includegraphics[width=0.9\columnwidth]{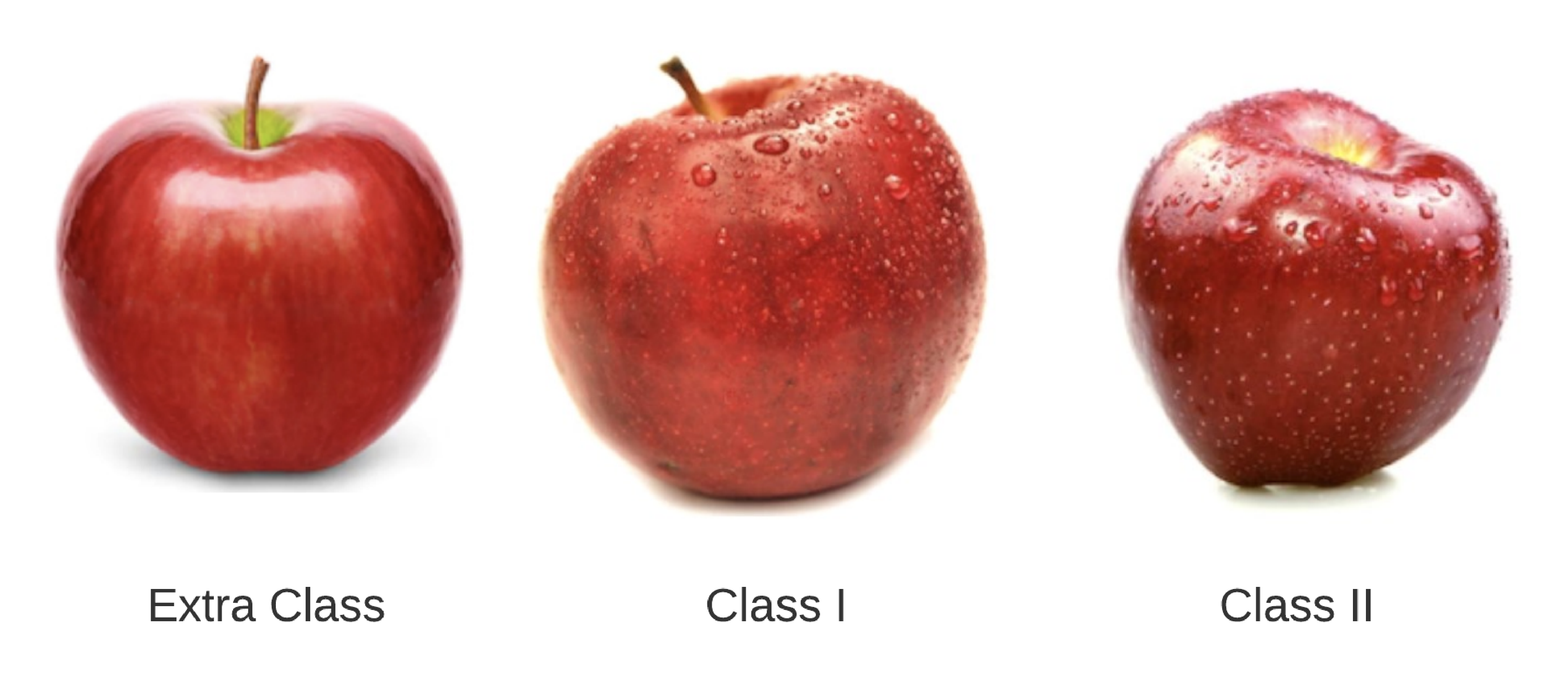}
  \caption{Apple Classification based on OECD standard.}
  \label{fig:applesComplete}
\end{figure}

According to~\cite{porat2018postharvest}, the classification of deformities in apples is reflected in the classes depicted in Figure \ref{fig:applesComplete}: Extra Class, Class 1, Class 2. The OECD stipulates that apples must have a symmetrical surface without evident deviations or bumps. This clearly describes the typical shape of an apple, which, during its development, should have a symmetrical form on both sides, as if it were a mirror image. When the apple lacks this symmetrical shape, it is considered deformed~\cite{chakrabarti2021cusp}.

\begin{figure}[t]
  \centering
   \includegraphics[width=0.9\columnwidth]{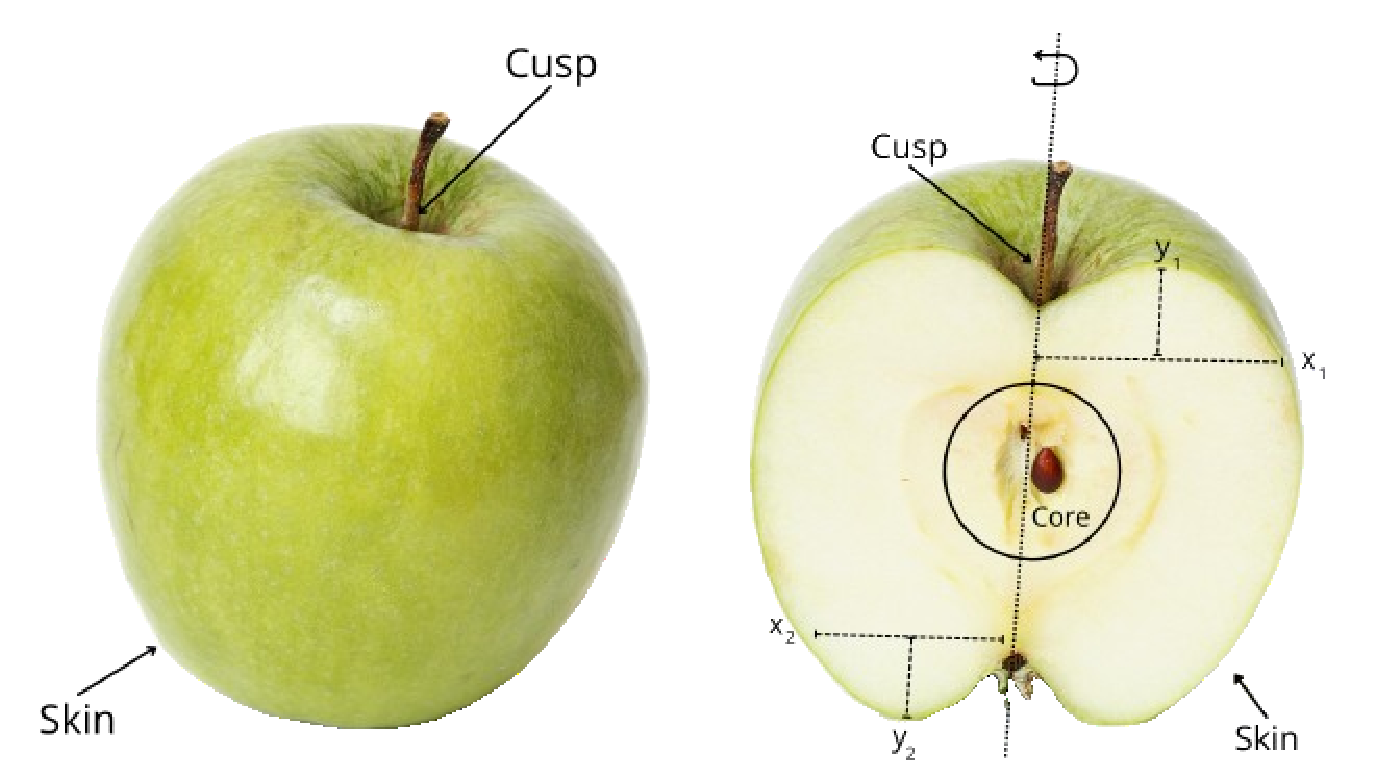}
  \caption{Apple Symmetry.}
  \label{fig:appleStudy}
\end{figure}

An apple whose distance between one end and its axis is similar to the distance between the axis and the other is considered symmetrical. As can be observed in Figure \ref{fig:appleStudy}, both sides of the "top" part of the apple grow similarly.

In the context of this research, all apples, regardless of their family, must have a symmetrical shape. Therefore, apples from different families can be included in the image dataset since they all contribute to the recognition of deformities in their symmetry.

\subsubsection{Mangoes Deformities}

\begin{figure}
  \centering
   \includegraphics[width=0.7\columnwidth]{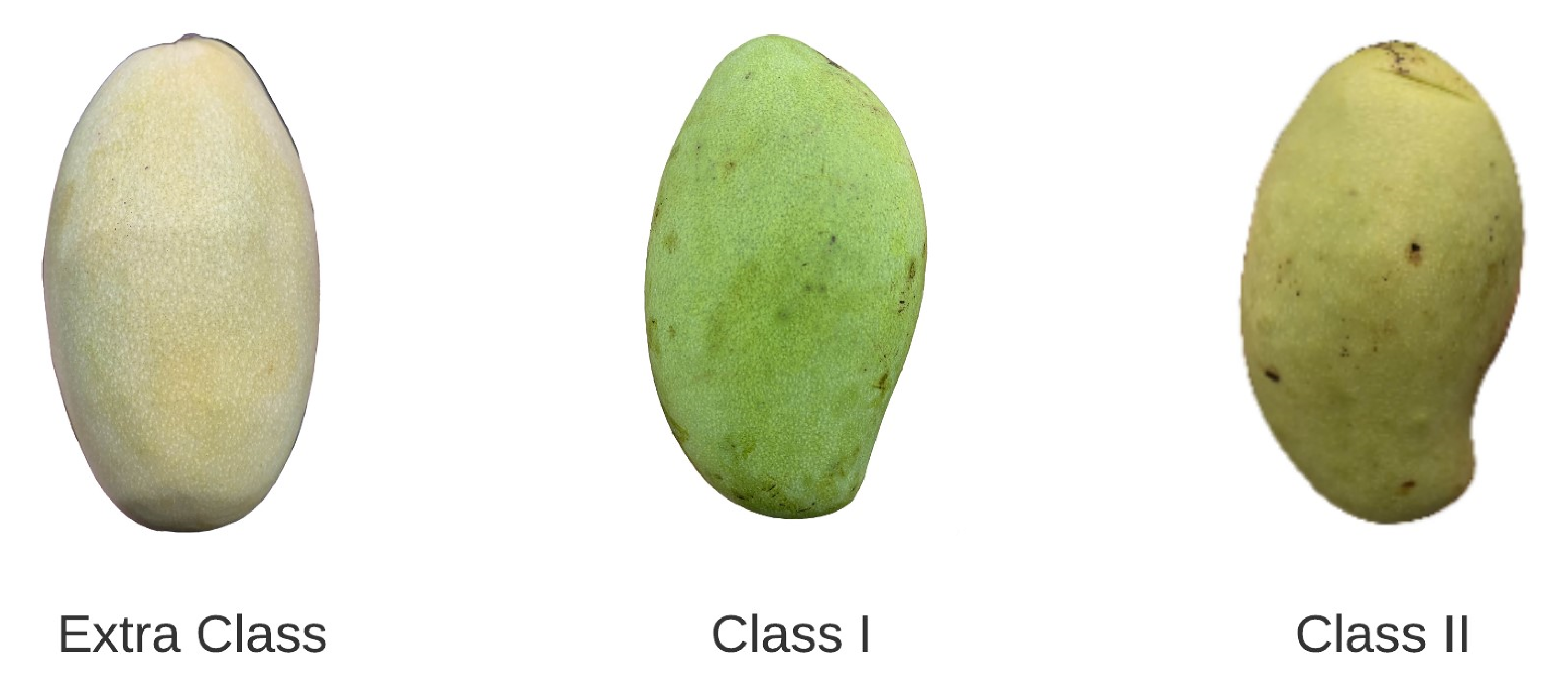}
  \caption{Mangoes Classification based on OECD standard.}
  \label{fig:mangoComplete}
\end{figure}

\begin{figure}
  \centering
   \includegraphics[width=0.7\columnwidth]{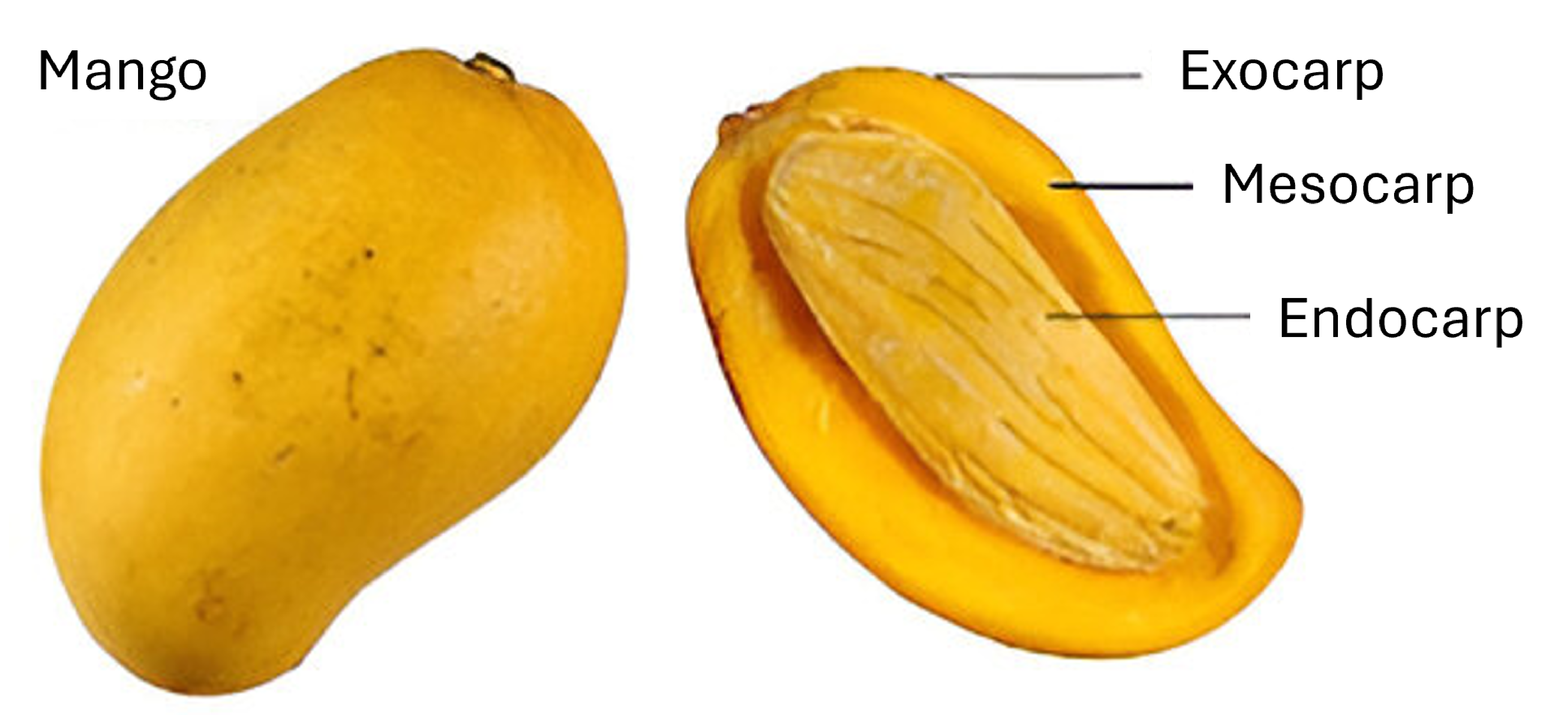}
  \caption{Mango with elongated oval shape.}
  \label{fig:mgFinal}
\end{figure}

According to~\cite{porat2018postharvest}, the classification of deformities in mangoes is reflected in the classes depicted in Figure \ref{fig:mangoComplete}: Extra Class, Class 1, Class 2. According to the parameters set by the OECD, mangoes should have an oval shape. However, there are different mango families, so we cannot guarantee that every mango will have a similar shape. This study will focus on the properties of mangoes from Mangifera Indica, specifically the Tommy Atkins variety, and its traditional shape is ovoid (see Fig. \ref{fig:mgFinal}).

\subsubsection{Strawberries Deformities}

\begin{figure}
  \centering
  \includegraphics[width=0.8\columnwidth]{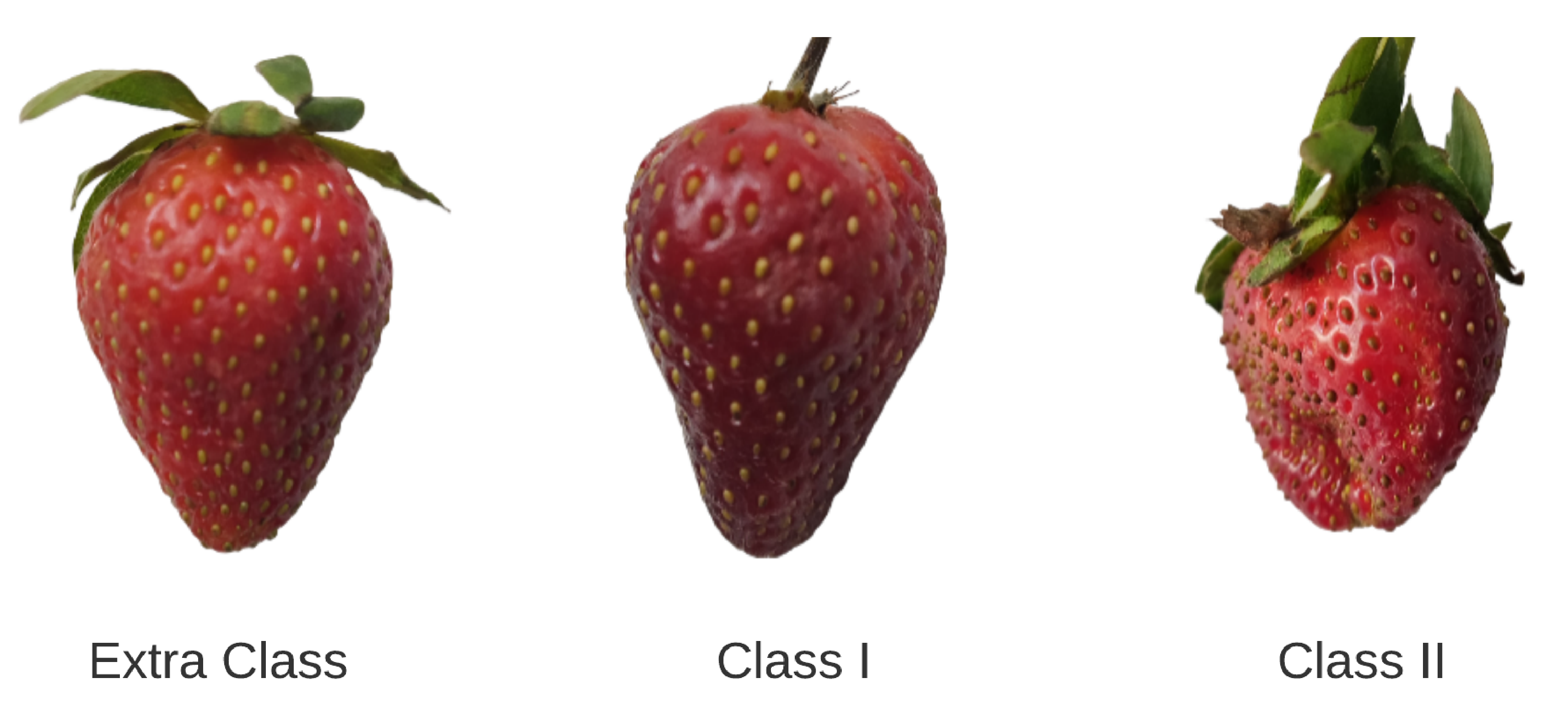}
  \caption{Strawberries Classification based on OECD standard.}
  \label{fig:strawberriesComplete}
\end{figure}

\begin{figure}
  \centering
   \includegraphics[width=0.7\columnwidth]{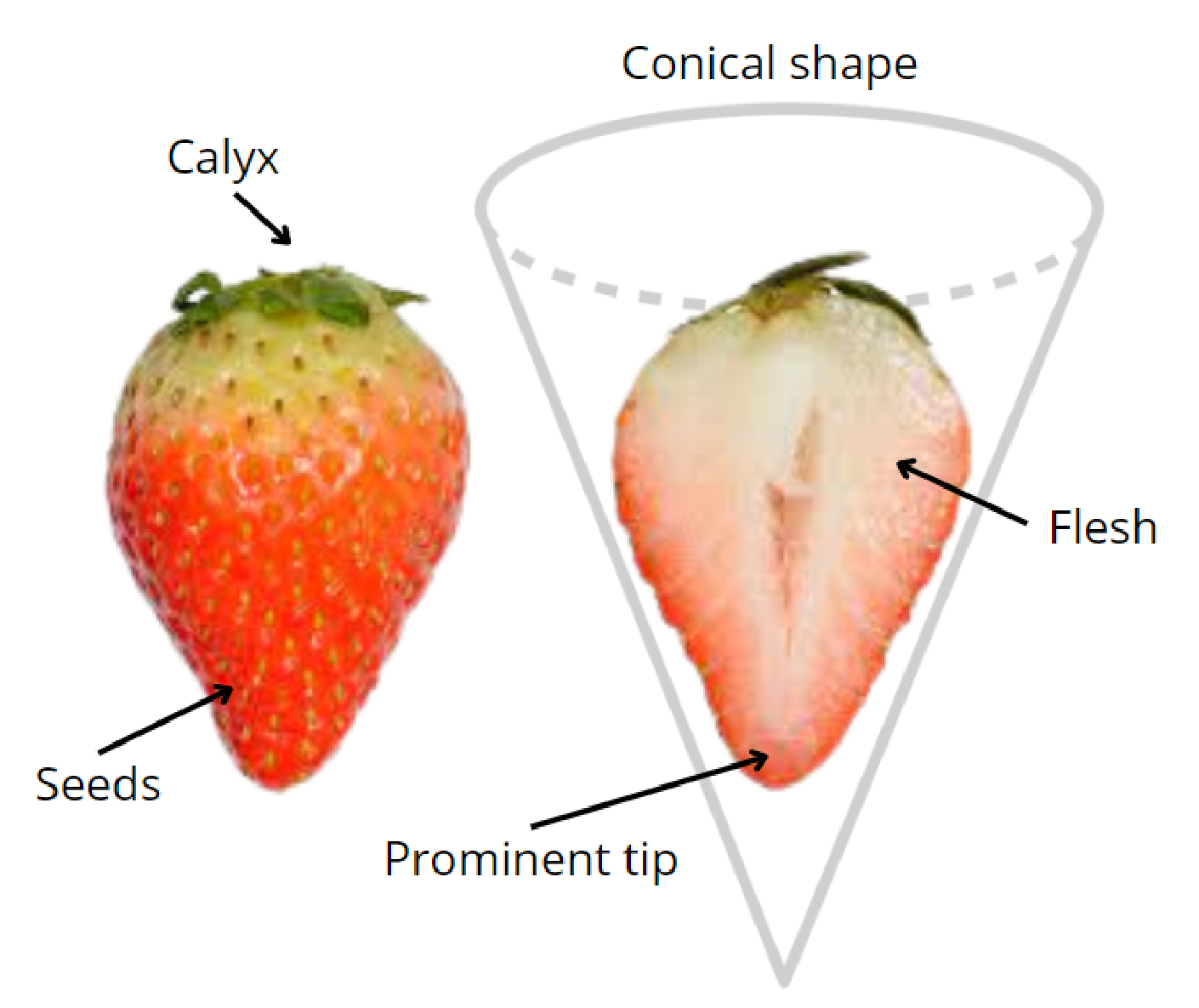}
  \caption{Strawberry Symmetry.}
  \label{fig:stFinal}
\end{figure}

According to~\cite{porat2018postharvest}, the classification of deformities in strawberries is reflected in the classes depicted in Figure \ref{fig:strawberriesComplete}: Extra Class, Class 1, Class 2. Just like with apples and mangoes, there are various species and families of strawberries. Fortunately for this study, strawberry varieties are distinguished by their size and flavor characteristics, but the traditional conical shape is maintained~\cite{liu2016flavor}. As a result, any strawberry family will be included in the image collection (see Fig. \ref{fig:stFinal}). A strawberry is considered symmetrical if its shape from the calyx to the tip is like an inverted cone.

\subsection{Generation of datasets}

When utilizing CNN models, it is essential to have an extensive database of images from which the model can extract meaningful features for each classification. However, having a large quantity of training images does not conclusively equate to obtaining a model with high accuracy. In~\cite{vujovic2021classification}, it was concluded that inadequate preprocessing of input images resulted in a model with low learning levels. From this perspective, the need to refine the dataset to ensure the quality of the input data for CNN models becomes evident.

Studies such as~\cite{wang2022grading},~\cite{sun2021improved}, and~\cite{garillos2021multimodal} add value to their image datasets by employing preprocessing techniques, which involve segmenting the object of interest, applying enhancement processes, or using images in different visible spectra.

When acquiring images, it is common to turn to public platforms, although at times it is crucial to maintain strict control over the input data, as discussed in~\cite{wang2022grading}. On the other hand, a proprietary acquisition process is often chosen, where the environment is prepared, and parameters such as luminosity and focus distance can be managed, among others. Furthermore, if the aim is to increase the number of training images, Data Augmentation techniques can be employed. The use of synthetic images is also an option. The work in~\cite{chuquimarca2023banana} utilizes the Unreal Engine game engine to create realistic images of bananas at different stages of ripeness. In \cite{pacheco2023fruit}, on the other hand, DALL-E mini, a Text-to-Image artificial intelligence, was used to generate images of apples and mangoes with various types of defects such as rot, scabs, and blemishes. In both studies, the use of this type of image contributed to improving the model's performance.

\subsection{Classification of deformities using CNN models}

The study in~\cite{hu2021infield} focused on identifying the shape and size of apples. A modified VGG16 model was used, which consisted of two fully connected layers (traditional in VGG architectures), and an additional four convolutional layers were added. As a result of the modification, the model achieved an accuracy of 99.04\%.

The study in~\cite{cao2021automated} proposes an apple size classification system using their own CNN, the LightNet model. The model's generalization capacity was verified as it was tested with images of zizanias, yielding similar results in model accuracy. The proposed method achieves an accuracy of 99.31\%, surpassing that obtained by the AlexNet, VGG11, ResNet18, MobileNet, MobileNetV2, and ShuffleNetV2 models.

Currently, there is a noticeable scarcity of published works on the classification of deformities in fruits using CNN models. This research gap highlights a significant opportunity for future investigations in the field. The application of deep learning techniques, such as CNNs, in the detection and classification of deformities in fruits represents a promising and relevant area, both for the agricultural industry and for the automation of quality inspection processes. Therefore, researchers are encouraged to explore and develop innovative approaches in this direction, aiming to enhance the efficiency and accuracy of fruit deformity classification.

\subsection{Architectures Topology}

Single-Input Architectures process a single type of input data at a time. This is the most common architecture and is used in a wide range of applications. Multi-Input Architectures can process two or more types of input data simultaneously. This allows the network to learn complex relationships between different types of data, such as RGB images and images in other spectrums. The architecture has multiple branches, each processing a type of input data, and then concatenates the features extracted from each branch to perform the final classification or prediction task~\cite{mesa2021multi,pipitsunthonsan2023palm}.

\section{Proposed Methodology}
\label{sec:methodology}

Throughout this section, the methodology employed in the development of this work is outlined. Additionally, it is important to highlight the contributions that this study offers to the scientific community:

\begin{itemize}
\item Three distinct datasets were developed for this study: Initially, a real dataset was compiled, sourced from public image repositories, and carefully refined according to international standards for shape classification. The second dataset, synthetic in nature, was generated using a blend of "text-to-image" and "image-to-image" tools, to produce high-fidelity images to enrich the training of CNN models. Lastly, a dataset comprising fruit silhouettes was created, wherein only the fruit shapes were meticulously extracted. Furthermore, the dataset generated by this study is made publicly available at the following link: (\url{https://github.com/xxxxxxx/Deformed-fruits.git})
\item The accuracies obtained by the three evaluated models: VGG16, MobileNetV2, and CIDIS are compared. It is important to highlight that the MobileNetV2 model has demonstrated superior performance compared to the other evaluated CNN models.
\item The contribution of this study lies in the evaluation of two different training approaches: Single-Input architecture, which allows for progressive and linear learning in contrast to Multi-Input architecture training, involving the simultaneous training of two models with different input approaches. In the case of Multi-Input architecture training, two branches are trained separately, one with RGB images and the other with silhouette images.
\end{itemize}

\subsection{Real Image Acquisition}

As with any machine learning model, the availability of a large number of images is essential for CNN models to effectively learn and generalize results. In this regard, it was considered that the appropriate number of images for each deformity category would be around 5,000; meaning a total of 20,000 images per fruit type were required to classify the 4 types of deformities. To compile this dataset, various online sources were explored, such as Kaggle, Github, and Mendeley. Initially, datasets already classified similarly to the established categories (Extra, Class I, Class II, and Unclassified) were sought. However, the search did not yield any public datasets with these specific classes, although datasets intended for other types of classifications were found. Faced with this situation, it was decided to adopt any available fruit image datasets, such as apples, mangoes, and strawberries, and subsequently classify the images according to the deformity class. Following this, the manual classification of each image obtained from public platforms was refined. Each image was categorized according to the accepted limits for each class. At the same time, the dataset was curated by removing fruit images that did not contribute significantly to the training process. This included images with non-uniform backgrounds that were difficult to remove, as well as images of fruits captured from the top or bottom perspectives.

At the end of the classification process, a significant imbalance was observed among the deformity categories, with the Extra Class having the highest number of images, in contrast to Class II and Ungraded. To address this issue and balance the image dataset, data augmentation techniques were applied, increasing the number of images by a factor of up to 8x for the classes that required it.

\subsection{Synthetic Image Acquisition}
\label{sec:synthetic_image}

The generation of synthetic images was carried out in multiple stages, aiming to reach 5,000 images per class for each type of fruit. The initial phase entailed generating near-perfect images for each fruit and category using Easy Diffusion tool version 2.5~\cite{brade2023promptify}. This tool facilitated the creation of diverse image variants corresponding to the categories from texts. However, at a certain point, it ceased to provide significant variations, or the generated images diverged from the expected outcome. Consequently, the decision was made to transition to generating images from existing ones.

Following the approach outlined in~\cite{somepalli2023diffusion}, small sets of images (between 200 and 800) were created, depending on the different shape scales for each category (Extra, Class I, Class II, Ungraded) and fruit, especially for those classes that were poorer, such as Class II and Ungraded. Models were then trained using LORA with AUTOMATIC 1111~\cite{hidalgo2023personalizing}. With the results obtained, the image generator Kohya~\cite{masrouri2024towards} was used. Unlike Easy Diffusion, this tool allows us to load models trained and generated by AUTOMATIC 1111. This enables us to combine the trained image model with text and generate additional variations beyond what the model learned through LORA. This approach allowed the completion of the datasets to reach 5,000 synthetic images for each class for each fruit.

\subsection{Model Pre-training Approach}

\begin{figure}
  \centering
   \includegraphics[width=0.5\columnwidth]{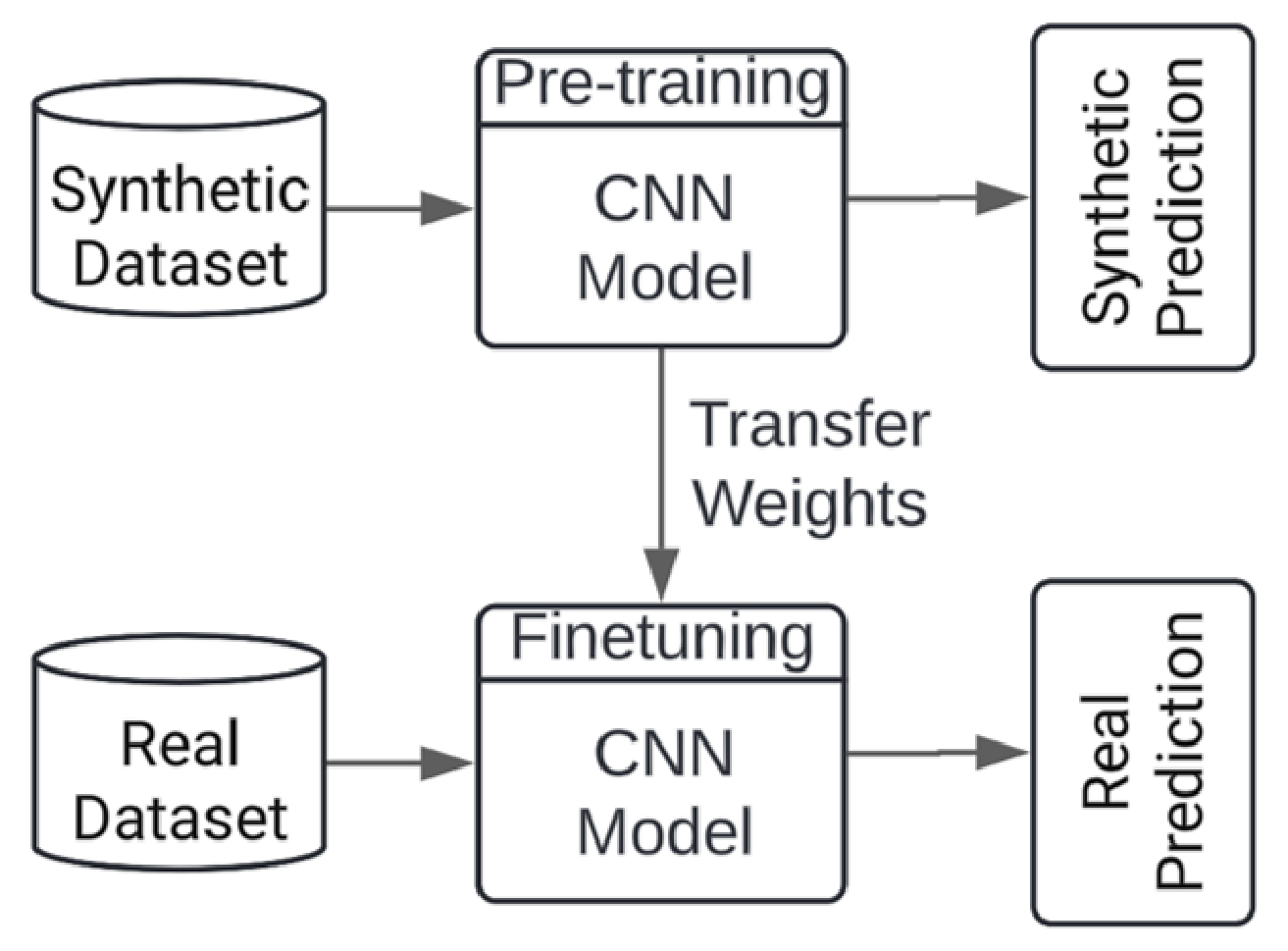}
  \caption{Model Pre-training Approach}
  \label{fig:imgTL}
\end{figure}

The training started from the CIDIS model~\cite{chuquimarca2023banana}. This model, being custom-made, does not have an initial weight matrix. Therefore, it was first trained with the synthetic image dataset. Once it learned the characteristics of the synthetic data, the learning was transferred to further training using the real image dataset. Figure \ref{fig:imgTL} illustrates the process graphically. In the Model Pre-training approach, we utilized the VGG16 and MobileNetV2 architectures with pre-trained weights, prepared for classification. Thus, in the Model Pre-training process of these architectures, only real images were employed, unlike the CIDIS model, which was used for the initial synthetic phase and subsequent Model Pre-training with real images.

\subsection{Multi-Input Network Approach}

\begin{figure}
  \centering
   \includegraphics[width=0.9\columnwidth]{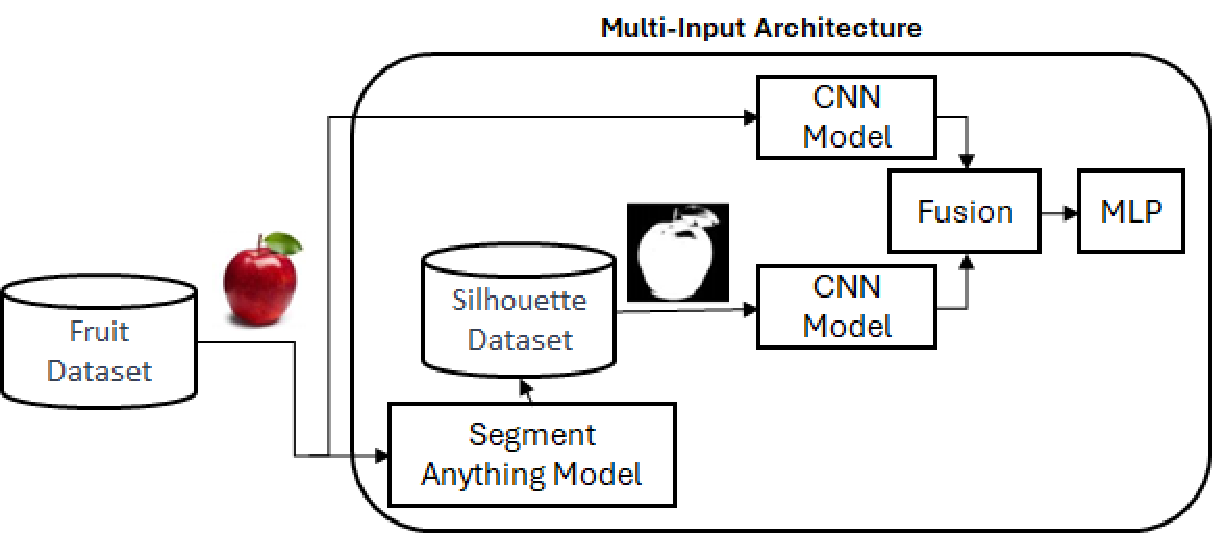}
  \caption{Multi-Input Network.}
  \label{fig:siameseNW}
\end{figure}

In the analysis of implementing a Multi-Input architecture, the focus lies primarily on evaluating the differences between the input images of each branch~\cite{mesa2021multi,dua2021multi,choudhary2023multi}. Subsequently, the result obtained from each branch is unified through a feature fusion process, and classification is carried out after a Multilayer Perceptron (MLP) layer. Given that this approach prioritizes the study of fruit shape, it was necessary to disregard features typically recognized by CNNs, such as color and texture in the images. Therefore, it was decided that the input images for one of the branches would be RGB images of the fruits (from the real and synthetic datasets), while the other branch would receive the silhouettes of the same fruits in those images, as depicted in Figure \ref{fig:siameseNW}.

For this approach, the CIDIS, MobileNetV2, and VGG16 models were used again, with the difference that the CIDIS model was trained without initial weights, while the MobileNetV2 and VGG16 models were initialized with the same weights found before training with the Model Pre-training approach.

\section{Experimental Results}
\label{sec:results}

In this section, we elaborate on the datasets of real and synthetic images, which were both acquired and generated to ensure high-quality data for evaluating CNN models. Subsequently, we provide a detailed presentation of the results obtained for each training approach, taking into account the various CNN models utilized.

After gathering the minimum required set of real fruit images (650), data augmentation techniques, including rotations, horizontal flips, and vertical flips, were employed to achieve the agreed-upon number of images for training, validation, and testing (80/10/10). The resulting image size was fixed at 224x224 pixels. In summary, a total of 20,000 images per fruit were obtained, evenly distributed among the four deformation categories (see Table \ref{tab:number_real}).

\begin{table}
\caption{Number of acquired and augmented real images.}
\label{tab:number_real}
\centering
\begin{tabularx}{\textwidth}{|l|X|X|X|X|}
\hline
\textbf{Fruits} &
  \multicolumn{4}{c|}{\textbf{Class}} \\
\cline{2-5}
&
  \textbf{Extra Class} &
  \textbf{First Class} &
  \textbf{Second Class} &
  \textbf{Out of Class} \\
\hline
Apples &
  \begin{tabular}{@{}l@{}}
  4050 (950)
  \end{tabular} &
  \begin{tabular}{@{}l@{}}
  1741 (3259)
  \end{tabular} &
  \begin{tabular}{@{}l@{}}
  651 (4349)
  \end{tabular} &
  \begin{tabular}{@{}l@{}}
  1961 (3039)
  \end{tabular} \\
\hline
Strawberry &
  \begin{tabular}{@{}l@{}}
  2782 (2218)
  \end{tabular} &
  \begin{tabular}{@{}l@{}}
  875 (4125)
  \end{tabular} &
  \begin{tabular}{@{}l@{}}
  759 (4241)
  \end{tabular} &
  \begin{tabular}{@{}l@{}}
  679 (4321)
  \end{tabular} \\
\hline
Mango &
  \begin{tabular}{@{}l@{}}
  4317 (683)
  \end{tabular} &
  \begin{tabular}{@{}l@{}}
  1889 (3111)
  \end{tabular} &
  \begin{tabular}{@{}l@{}}
  655 (4345)
  \end{tabular} &
  \begin{tabular}{@{}l@{}}
  734 (4266)
  \end{tabular} \\
\hline
\end{tabularx}
\end{table}

On the other hand, given the scarcity of suitable real images for fruit deformity analysis, synthetic datasets were generated leveraging the capabilities provided by the Stable Diffusion tool, as well as the technologies mentioned in Section~\ref{sec:synthetic_image}. These tools produced high-quality synthetic images with dimensions of 512x512 pixels. In many cases, images from the real dataset were used as a basis for generating similar synthetic images. Subsequently, data augmentation techniques were employed to rotate and flip the images, aiming to introduce variability into the dataset. Similar to the real image set, a total of 20,000 images per fruit were obtained, evenly distributed among the deformity categories.

\begin{figure}
  \centering
   \includegraphics[width=0.5\columnwidth]{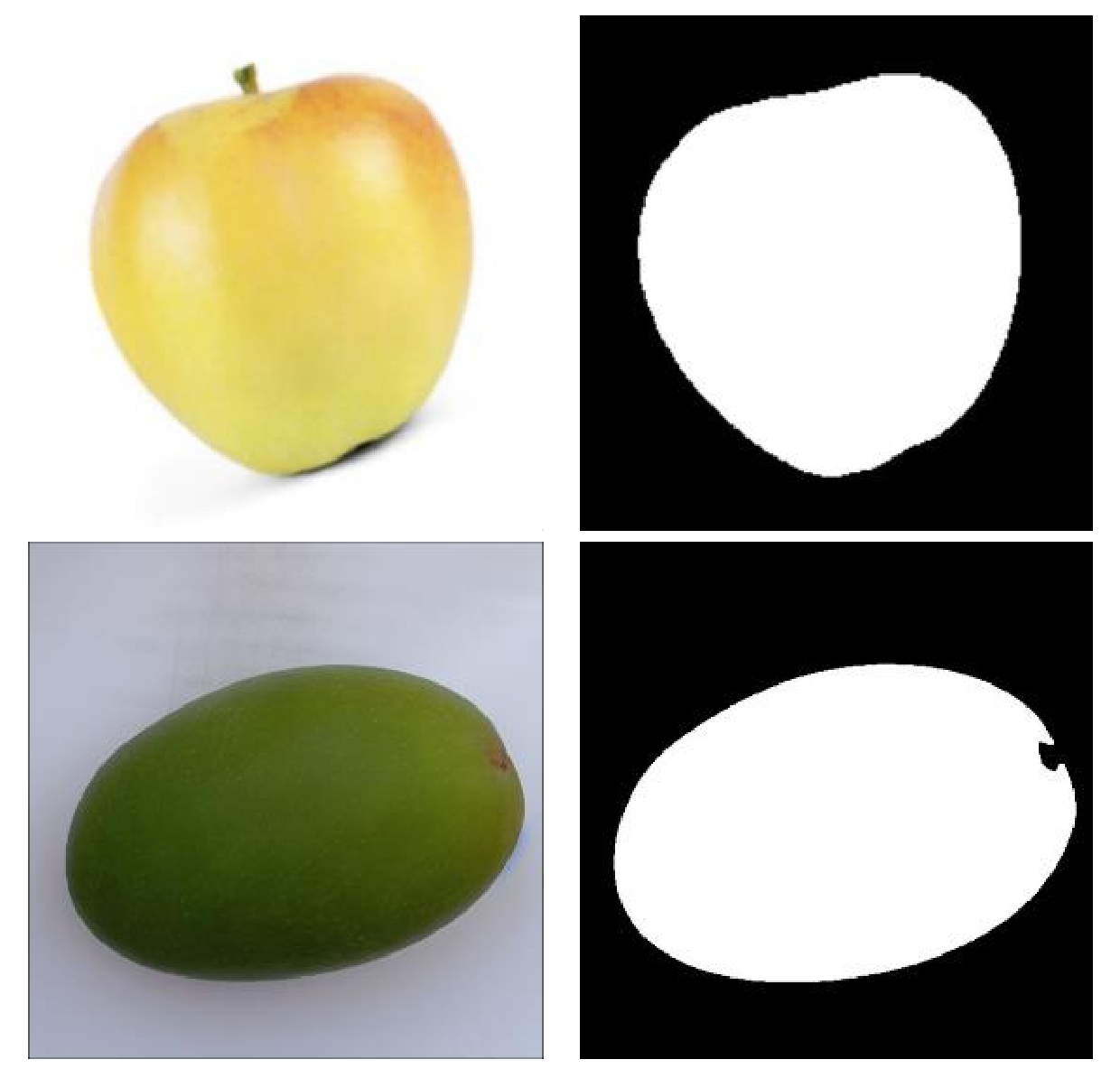}
  \caption{Obtaining the Silhouette of Each Fruit}
  \label{fig:siluetasMApp}
\end{figure}

Furthermore, silhouette images were prepared, which are closely linked to the shape of an object. In our case, fruit analysis revolves around its shape. Therefore, it is crucial to train the models using fruit silhouettes. For the automatic generation of these silhouette images, the Segment Anything Model (SAM) tool was utilized. The generation of fruit silhouette images was successful in most cases (see Fig. \ref{fig:siluetasMApp}). For fruit images with incorrect silhouettes, individual processing was applied, or they were excluded entirely, to obtain a refined and high-quality dataset.

Once the datasets were prepared, hyperparameters were set for training each CNN model. Each training was conducted with a different combination based on the obtained results. The use of various optimizers, such as Adam, Nadam, and RMSProp, along with variable learning rates, is noted. Typically, trainings were configured to run until epoch 30, and the learned weights were restored from the epoch with the best performance.

\subsection{Results with Model Pre-training approach}

When training and evaluating the CNN models using only real data, the results shown in Table \ref{tab:tab22} were obtained. These results provide a detailed assessment of the models' performance, including validation accuracy, validation loss, precision, recall, F1-score, and test accuracy. Table \ref{tab:tab22} offers a thorough comparison of these performance metrics across different fruits and model architectures, providing a clear understanding of each approach's effectiveness in image classification. However, the results shown in Table \ref{tab:tab33} are those obtained with the Model Pre-training approach, which achieves better results than the models trained only with real images. We based our selection on the model with the highest accuracy and the lowest loss.

\begin{table}
\caption{Training Results with CIDIS, MobileNetV2, and VGG16, Model Training Approach using only Real Data.}
\label{tab:tab22}
\centering
\begin{tabular}{|l|c|c|c|c|c|c|c|}
\hline
\textbf{Fruit} &
  \textbf{Model} &
  \textbf{Val Accuracy} &
  \textbf{Val Loss} & 
  \textbf{Precision} &
  \textbf{Recall} &
  \textbf{F1-score} &
  \textbf{Test Accuracy} \\
\hline
Apple &
  \begin{tabular}[c]{@{}c@{}}
  CIDIS \\
  \textbf{MobileNetV2} \\
  VGG16
  \end{tabular} &
  \begin{tabular}[c]{@{}c@{}}
  0.6815 \\
  \textbf{0.7848} \\
  0.7747
  \end{tabular} &
  \begin{tabular}[c]{@{}c@{}}
  0.8491 \\
  \textbf{0.5851} \\
  0.6073
  \end{tabular} &
  \begin{tabular}[c]{@{}c@{}}
  0.6100 \\
  \textbf{0.6100} \\
  0.5600
  \end{tabular} &
  \begin{tabular}[c]{@{}c@{}}
  0.5900 \\
  \textbf{0.6100} \\
  0.5700
  \end{tabular} &
  \begin{tabular}[c]{@{}c@{}}
  0.5500 \\
  \textbf{0.5700} \\
  0.5200
  \end{tabular} &
  \begin{tabular}[c]{@{}c@{}}
  0.5915 \\
  \textbf{0.6095} \\
  0.5685
  \end{tabular} \\
\hline
Mango &
  \begin{tabular}[c]{@{}c@{}}
  CIDIS \\
  \textbf{MobileNetV2} \\
  VGG16
  \end{tabular} &
  \begin{tabular}[c]{@{}c@{}}
  0.5155 \\
  \textbf{0.5155} \\
  0.5746
  \end{tabular} &
  \begin{tabular}[c]{@{}c@{}}
  1.2260 \\
  \textbf{0.9612} \\
  1.0673
  \end{tabular} &
  \begin{tabular}[c]{@{}c@{}}
  0.4400 \\
  \textbf{0.6700} \\
  0.4900
  \end{tabular} &
  \begin{tabular}[c]{@{}c@{}}
  0.4100 \\
  \textbf{0.6700} \\
  0.4800
  \end{tabular} &
  \begin{tabular}[c]{@{}c@{}}
  0.4100 \\
  \textbf{0.6700} \\
  0.4800
  \end{tabular} &
  \begin{tabular}[c]{@{}c@{}}
  0.4400 \\
  \textbf{0.6710} \\
  0.4770
  \end{tabular} \\
\hline
Strawberry &
  \begin{tabular}[c]{@{}c@{}}
  CIDIS \\
  \textbf{MobileNetV2} \\
  VGG16
  \end{tabular} &
  \begin{tabular}[c]{@{}c@{}}
  0.5601 \\
  \textbf{0.6971} \\
  0.6694
  \end{tabular} &
  \begin{tabular}[c]{@{}c@{}}
  1.3034 \\
  \textbf{1.2005} \\
  1.0141
  \end{tabular} &
  \begin{tabular}[c]{@{}c@{}}
  0.4900 \\
  \textbf{0.7000} \\
  0.6300
  \end{tabular} &
  \begin{tabular}[c]{@{}c@{}}
  0.4100 \\
  \textbf{0.7000} \\
  0.6100
  \end{tabular} &
  \begin{tabular}[c]{@{}c@{}}
  0.4000 \\
  \textbf{0.7000} \\
  0.6000
  \end{tabular} &
  \begin{tabular}[c]{@{}c@{}}
  0.4110 \\
  \textbf{0.6980} \\
  0.6100
  \end{tabular} \\
\hline
\end{tabular}
\end{table}

\begin{table}
\caption{Training Results with CIDIS, MobileNetV2, and VGG16, Model Pre-training approach.}
\label{tab:tab33}
\centering
\begin{tabular}{|l|c|c|c|c|c|c|c|}
\hline
\textbf{Fruit} &
  \textbf{Model} &
  \textbf{Val Accuracy} &
  \textbf{Val Loss} & 
  \textbf{Precision} &
  \textbf{Recall} &
  \textbf{F1-score} &
  \textbf{Test Accuracy} \\
\hline
Apple &
  \begin{tabular}[c]{@{}c@{}}
  CIDIS \\
  \textbf{MobileNetV2} \\
  VGG16
  \end{tabular} &
  \begin{tabular}[c]{@{}c@{}}
  0.8956 \\
  \textbf{0.9325} \\
  0.9050
  \end{tabular} &
  \begin{tabular}[c]{@{}c@{}}
  0.3645 \\
  \textbf{0.2087} \\
  0.2511
  \end{tabular} &
  \begin{tabular}[c]{@{}c@{}}
  0.8900 \\
  \textbf{0.9100} \\
  0.9000
  \end{tabular} &
  \begin{tabular}[c]{@{}c@{}}
  0.8800 \\
  \textbf{0.9000} \\
  0.8900
  \end{tabular} &
  \begin{tabular}[c]{@{}c@{}}
  0.8800 \\
  \textbf{0.9000} \\
  0.8900
  \end{tabular} &
  \begin{tabular}[c]{@{}c@{}}
  0.8806 \\
  \textbf{0.9044} \\
  0.8931
  \end{tabular} \\
\hline
Mango &
  \begin{tabular}[c]{@{}c@{}}
  CIDIS \\
  \textbf{MobileNetV2} \\
  VGG16
  \end{tabular} &
  \begin{tabular}[c]{@{}c@{}}
  0.08679 \\
  \textbf{0.9420} \\
  0.9022
  \end{tabular} &
  \begin{tabular}[c]{@{}c@{}}
  0.3819 \\
  \textbf{0.1594} \\
  0.2290
  \end{tabular} &
  \begin{tabular}[c]{@{}c@{}}
  0.8700 \\
  \textbf{0.9400} \\
  0.9600
  \end{tabular} &
  \begin{tabular}[c]{@{}c@{}}
  0.8700 \\
  \textbf{0.9400} \\
  0.9600
  \end{tabular} &
  \begin{tabular}[c]{@{}c@{}}
  0.8700 \\
  \textbf{0.9400} \\
  0.9600
  \end{tabular} &
  \begin{tabular}[c]{@{}c@{}}
  0.8694 \\
  \textbf{0.9425} \\
  0.9604
  \end{tabular} \\
\hline
Strawberry &
  \begin{tabular}[c]{@{}c@{}}
  CIDIS \\
  \textbf{MobileNetV2} \\
  VGG16
  \end{tabular} &
  \begin{tabular}[c]{@{}c@{}}
  0.8010 \\
  \textbf{0.9189} \\
  0.8962
  \end{tabular} &
  \begin{tabular}[c]{@{}c@{}}
  0.5258 \\
  \textbf{0.1945} \\
  0.2646
  \end{tabular} &
  \begin{tabular}[c]{@{}c@{}}
  0.8200 \\
  \textbf{0.9200} \\
  0.9100
  \end{tabular} &
  \begin{tabular}[c]{@{}c@{}}
  0.8100 \\
  \textbf{0.9200} \\
  0.9100
  \end{tabular} &
  \begin{tabular}[c]{@{}c@{}}
  0.8200 \\
  \textbf{0.9200} \\
  0.9100
  \end{tabular} &
  \begin{tabular}[c]{@{}c@{}}
  0.8135 \\
  \textbf{0.9220} \\
  0.9050
  \end{tabular} \\
\hline
\end{tabular}
\end{table}

In this way, it is determined that the best architecture for this type of training has been MobileNet for all fruits. It is worth mentioning that in the case of mango, VGG16 had the highest precision; however, the loss was very high. Therefore, we consider that a better result can be achieved with MobileNet.

\subsection{Results with Multi-Input Network Approach}

We relied on studies that validated the results of Multi-Input networks for extracting features between two similar images. In our approach, we feed one branch with an RGB image and the other with the silhouette of the same image. This process effectively identifies similarities or differences in the shape of the object in the image, in our case, the shape of the fruit. The Multi-Input network approach was employed with the complete dataset of real and synthetic images, resulting in a larger volume of images than Model Pre-training. All models obtained with this approach exhibited an accuracy above 90\%, with the MobileNetV2 model achieving the highest accuracy for apples and strawberries, whereas for mangoes, the VGG16 model was superior (see Table~\ref{tab:tab34}).

\begin{table}
\caption{Training Results with CIDIS, MobileNetV2, and VGG16, Multi-Input Network Approach}
\label{tab:tab34}
\centering
\begin{tabular}{|l|c|c|c|c|c|c|c|}
\hline
\textbf{Fruit} &
  \textbf{Model} &
  \textbf{Val Accuracy} &
  \textbf{Val Loss} & 
  \textbf{Precision} &
  \textbf{Recall} &
  \textbf{F1-score} &
  \textbf{Test Accuracy} \\
\hline
Apple &
  \begin{tabular}[c]{@{}c@{}}
  CIDIS \\
  \textbf{MobileNetV2} \\
  VGG16
  \end{tabular} &
  \begin{tabular}[c]{@{}c@{}}
  0.7626 \\
  \textbf{0.8068} \\
  0.8060
  \end{tabular} &
  \begin{tabular}[c]{@{}c@{}}
  1.6818 \\
  \textbf{2.2276} \\
  1.9643
  \end{tabular} &
  \begin{tabular}[c]{@{}c@{}}
  0.8639 \\
  \textbf{0.9257} \\
  0.8966
  \end{tabular} &
  \begin{tabular}[c]{@{}c@{}}
  0.8366 \\
  \textbf{0.9158} \\
  0.8920
  \end{tabular} &
  \begin{tabular}[c]{@{}c@{}}
  0.8107 \\
  \textbf{0.9161} \\
  0.8832
  \end{tabular} &
  \begin{tabular}[c]{@{}c@{}}
  0.8249 \\
  \textbf{0.9257} \\
  0.8844
  \end{tabular} \\
\hline
Mango &
  \begin{tabular}[c]{@{}c@{}}
  CIDIS \\
  MobileNetV2 \\
  \textbf{VGG16}
  \end{tabular} &
  \begin{tabular}[c]{@{}c@{}}
  0.9234 \\
  0.9440 \\
  \textbf{0.9501}
  \end{tabular} &
  \begin{tabular}[c]{@{}c@{}}
  0.2577 \\
  0.1505 \\
  \textbf{0.1696}
  \end{tabular} &
  \begin{tabular}[c]{@{}c@{}}
  0.9237 \\
  0.9394 \\
  \textbf{0.9551}
  \end{tabular} &
  \begin{tabular}[c]{@{}c@{}}
  0.9232 \\
  0.9393 \\
  \textbf{0.9539}
  \end{tabular} &
  \begin{tabular}[c]{@{}c@{}}
  0.9233 \\
  0.9390 \\
  \textbf{0.9541}
  \end{tabular} &
  \begin{tabular}[c]{@{}c@{}}
  0.9232 \\
  0.9393 \\
  \textbf{0.9551}
  \end{tabular} \\
\hline
Strawberry &
  \begin{tabular}[c]{@{}c@{}}
  CIDIS \\
  \textbf{MobileNetV2} \\
  VGG16
  \end{tabular} &
  \begin{tabular}[c]{@{}c@{}}
  0.9188 \\
  \textbf{0.9197} \\
  0.8769
  \end{tabular} &
  \begin{tabular}[c]{@{}c@{}}
  0.2496 \\
  \textbf{0.2330} \\
  0.4013
  \end{tabular} &
  \begin{tabular}[c]{@{}c@{}}
  0.9145 \\
  \textbf{0.9455} \\
  0.9623
  \end{tabular} &
  \begin{tabular}[c]{@{}c@{}}
  0.9171 \\
  \textbf{0.9228} \\
  0.9204
  \end{tabular} &
  \begin{tabular}[c]{@{}c@{}}
  0.9154 \\
  \textbf{0.9213} \\
  0.9190
  \end{tabular} &
  \begin{tabular}[c]{@{}c@{}}
  0.9171 \\
  \textbf{0.9394} \\
  0.9204
  \end{tabular} \\
\hline
\end{tabular}
\end{table}

In this case, it is evident that the MobileNetV2 model with the Multi-Input architecture yielded the best results for apples and strawberries, with an accuracy of 92\% and 93\%, respectively. However, the VGG16 model proved superior for mangoes with an accuracy of 95\%. Consequently, the most favorable results for the Multi-Input approach are summarized in Table~\ref{tab:tab35}.

\begin{table}
\caption{Results Comparison}
\label{tab:tab35}
\centering
\begin{tabular}{|l|l|c|}
\hline
\textbf{Fruits}                      & \textbf{Models}                                                               & \textbf{Accuracy} \\ \hline
\multirow{2}{*}{Apple}      & \begin{tabular}[c]{@{}c@{}}MobileNetV2 \\ (Single-Input)\end{tabular} & 0.9044   \\
                            & \begin{tabular}[c]{@{}c@{}}\textbf{MobileNetV2} \\ \textbf{(Multi-Input)}\end{tabular}      & \textbf{0.9257}   \\ \hline
\multirow{2}{*}{Mango}      & \begin{tabular}[c]{@{}c@{}}MobileNetV2 \\ (Single-Input)\end{tabular} & 0.9425   \\
                            & \begin{tabular}[c]{@{}c@{}}\textbf{VGG16} \\ \textbf{(Multi-Input)}\end{tabular}            & \textbf{0.9551}   \\ \hline
\multirow{2}{*}{Strawberry} & \begin{tabular}[c]{@{}c@{}}MobileNetV2 \\ (Single-Input)\end{tabular} & 0.9220   \\
                            & \begin{tabular}[c]{@{}c@{}}\textbf{MobileNetV2} \\ \textbf{(Multi-Input)}\end{tabular}      & \textbf{0.9394}   \\ \hline
\end{tabular}
\end{table}

\section{Conclusions}
\label{sec:conclusions}

A public dataset has been generated, presenting a wide range of images, including both real and synthetic images of apples, mangoes, and strawberries, thus addressing their morphological diversities. This dataset is evenly divided for each fruit into 5,000 samples for each of the four defined classes: Extra Class, First Class, Second Class, and Out of Class. Furthermore, it is supplemented with an additional set of 20,000 silhouette images generated from each RGB image, providing a more comprehensive and detailed representation of the morphological characteristics of the apples in the dataset. For this work, a dataset was obtained that enabled successful training. AI data generation tools such as Stable Diffusion played a fundamental role in achieving this accomplishment. Analyzing the shape of fruits involves being very careful with the images used. The results obtained show that training with binary images where only shape details are present significantly aids learning, as is the case with the Multi-Input Network approach. To achieve the goal of having a good dataset, it is essential to obtain as many images as possible that are easy to segment, such as images with a consistent background. Evaluating different architectures such as MobileNetV2, VGG16, and the one proposed by CIDIS has helped us understand that for this type of task, it is more advantageous to use lightweight architectures like MobileNetV2, as it yielded the best accuracy results. Both approaches, Model Pre-training, and the Multi-Input Network, have yielded good results in deformity classification. However, the Multi-Input Network approach stood out for its efficiency, grounded in the capability of the branches corresponding to RGB images and silhouette images to capture distinctive characteristics of fruit deformities. In the case of the branch processing RGB images, it benefited from detailed information regarding edges, textures, color patterns, and specific shapes, all of which are fundamental elements for the precise identification of deformities. Conversely, the branch associated with silhouette images allowed for the consideration of aspects such as shape, contours, edges, and pixel distribution within the silhouette, thereby enriching the representation of the structural characteristics of the fruits. For future work, a comparison with ViT and DINO ViT models is proposed, as they have demonstrated high efficacy in image classification tasks, to evaluate their performance in classifying fruit deformities compared to lightweight architectures like MobileNetV2. Additionally, it is essential to improve the quality of the dataset using advanced artificial intelligence tools to manage, clean, and select visual data quickly and at scale. These enhancements will ensure a more robust foundation for model training and evaluation, enhancing their ability to accurately and efficiently identify deformities.

\section*{Acknowledgment}
This work has been partially supported by the ESPOL-CIDIS-11-2022 project.

%
%
%
%


\begin{thebibliography}{8}

\bibitem{olorunfemi2021post}
Olorunfemi, B.J., Kayode, S.E.: Post-harvest loss and grain storage technology-a review. Turkish Journal of Agriculture-Food Science and Technology \textbf{9}(1), 75--83 (2021)

\bibitem{shewfelt2022challenges}
Shewfelt, R.L., Prussia, S.E.: Challenges in handling fresh fruits and vegetables. In: Postharvest Handling, pp. 167--186. Elsevier (2022)

\bibitem{vetrekar2015non}
Vetrekar, N.T., Gad, R.S., Fernandes, I., Parab, J.S., Desai, A.R., Pawar, J.D., Naik, G.M., Umapathy, S.: Non-invasive hyperspectral imaging approach for fruit quality control application and classification: case study of apple, chikoo, guava fruits. Journal of Food Science and Technology \textbf{52}, 6978--6989 (2015)

\bibitem{chuquimarca2023banana}
Chuquimarca, L.E., Vintimilla, B.X., Velastin, S.A.: Banana Ripeness Level Classification Using a Simple CNN Model Trained with Real and Synthetic Datasets. In: VISIGRAPP (5: VISAPP), pp. 536--543 (2023)

\bibitem{behera2020identification}
Behera, S.K., Rath, A.K., Mahapatra, A., Sethy, P.K.: Identification, classification \& grading of fruits using machine learning \& computer intelligence: a review. Journal of Ambient Intelligence and Humanized Computing, 1--11 (2020)

\bibitem{lidror1990improving}
Lidror, A., Prussia, S.E.: Improving quality assurance techniques for producing and handling agricultural crops. Journal of Food Quality \textbf{13}(3), 171--184 (1990)

\bibitem{huang2022us}
Huang, K.M., Guan, Z., Hammami, A.M.: The US fresh fruit and vegetable industry: An overview of production and trade. Agriculture \textbf{12}(10), 1719 (2022)

\bibitem{zhou2015food}
Zhou, J.H., Kai, L., Liang, Q.: Food safety controls in different governance structures in China's vegetable and fruit industry. Journal of Integrative Agriculture \textbf{14}(11), 2189--2202 (2015)

\bibitem{wang2022grading}
Wang, J., Huo, Y., Wang, Y., Zhao, H., Li, K., Liu, L., Shi, Y.: Grading detection of “Red Fuji” apple in Luochuan based on machine vision and near-infrared spectroscopy. Plos one \textbf{17}(8), e0271352 (2022)

\bibitem{hu2021infield}
Hu, G., Zhang, E., Zhou, J., Zhao, J., Gao, Z., Sugirbay, A., Jin, H., Zhang, S., Chen, J.: Infield apple detection and grading based on multi-feature fusion. Horticulturae \textbf{7}(9), 276 (2021)

\bibitem{porat2018postharvest}
Porat, R., Lichter, A., Terry, L.A., Harker, R., Buzby, J.: Postharvest losses of fruit and vegetables during retail and in consumers’ homes: Quantifications, causes, and means of prevention. Postharvest biology and technology \textbf{139}, 135--149 (2018)

\bibitem{chakrabarti2021cusp}
Chakrabarti, A., Michaels, T.C., Yin, S., Sun, E., Mahadevan, L.: The cusp of an apple. Nature Physics \textbf{17}(10), 1125--1129 (2021)

\bibitem{liu2016flavor}
Liu, L., Ji, M.L., Chen, M., Sun, M.Y., Fu, X.L., Li, L., Gao, D.S., Zhu, C.Y.: The flavor and nutritional characteristic of four strawberry varieties cultured in soilless system. Food Science \& Nutrition \textbf{4}(6), 858--868 (2016)

\bibitem{vujovic2021classification}
Vujović, Ž., et al.: Classification model evaluation metrics. International Journal of Advanced Computer Science and Applications \textbf{12}(6), 599--606 (2021)

\bibitem{sun2021improved}
Sun, L., Liang, K., Song, Y., Wang, Y.: An improved CNN-based apple appearance quality classification method with small samples. Ieee Access \textbf{9}, 68054--68065 (2021)

\bibitem{garillos2021multimodal}
Garillos-Manliguez, C.A., Chiang, J.Y.: Multimodal Deep Learning via Late Fusion for Non-Destructive Papaya Fruit Maturity Classification. In: 2021 18th International Conference on Electrical Engineering, Computing Science and Automatic Control (CCE), pp. 1--6. IEEE (2021)

\bibitem{pacheco2023fruit}
Pacheco, R., González, P., Chuquimarca, L.E., Vintimilla, B.X., Velastin, S.A.: Fruit Defect Detection Using CNN Models with Real and Virtual Data. In: VISIGRAPP (4: VISAPP), pp. 272--279 (2023)

\bibitem{cao2021automated}
Cao, J., Sun, T., Zhang, W., Zhong, M., Huang, B., Zhou, G., Chai, X.: An automated zizania quality grading method based on deep classification model. Computers and Electronics in Agriculture \textbf{183}, 106004 (2021)

\bibitem{mesa2021multi}
Mesa, Armacheska Rivero, and John Y Chiang. "Multi-input deep learning model with RGB and hyperspectral imaging for banana grading." \textit{Agriculture} 11.8 (2021): 687.

\bibitem{pipitsunthonsan2023palm}Pipitsunthonsan, P., Pan, L., Peng, S., Khaorapapong, T., Nakasathien, S., Channumsin, S. \& Chongcheawchamnan, M. Palm bunch grading technique using a multi-input and multi-label convolutional neural network. {\em Computers And Electronics In Agriculture}. \textbf{210} pp. 107864 (2023)

\bibitem{brade2023promptify}
Brade, S., Wang, B., Sousa, M., Oore, S., Grossman, T.: Promptify: Text-to-image generation through interactive prompt exploration with large language models. In: Proceedings of the 36th Annual ACM Symposium on User Interface Software and Technology, pp. 1--14 (2023)

\bibitem{somepalli2023diffusion}
Somepalli, G., Singla, V., Goldblum, M., Geiping, J., Goldstein, T.: Diffusion art or digital forgery? investigating data replication in diffusion models. In: Proceedings of the IEEE/CVF Conference on Computer Vision and Pattern Recognition, pp. 6048--6058 (2023)

\bibitem{hidalgo2023personalizing}
Hidalgo, R., Salah, N., Chandra Jetty, R., Jetty, A., Varde, A.S.: Personalizing text-to-image diffusion models by fine-tuning classification for AI applications. In: Intelligent Systems Conference, pp. 642--658. Springer (2023)

\bibitem{masrouri2024towards}
Masrouri, M., Qin, Z.: Towards data-efficient mechanical design of bicontinuous composites using generative AI. Theoretical and Applied Mechanics Letters \textbf{14}(1), 100492 (2024)

\bibitem{dua2021multi}
Dua, Nidhi, Shiva Nand Singh, and Vijay Bhaskar Semwal. "Multi-input CNN-GRU based human activity recognition using wearable sensors." \textit{Computing} 103.7 (2021): 1461--1478.

\bibitem{choudhary2023multi}
Choudhary, Anurag, Rismaya Kumar Mishra, Shahab Fatima, and BK Panigrahi. "Multi-input CNN based vibro-acoustic fusion for accurate fault diagnosis of induction motor." \textit{Engineering Applications of Artificial Intelligence} 120 (2023): 105872.




\end{thebibliography}


\end{document}